%% file: latex/acl_latex.tex
\documentclass[11pt]{article}

\usepackage[final]{acl}

\usepackage{times}
\usepackage{latexsym}

\usepackage[T1]{fontenc}

\usepackage[utf8]{inputenc}

\usepackage{microtype}

\usepackage{inconsolata}

\usepackage{graphicx}

\title{Towards Simulating Social Media Users with LLMs:\\ Evaluating the Operational Validity of Conditioned Comment Prediction}

\author{
  \textbf{Nils Schwager}$^1$, \textbf{Simon Münker}$^1$, \textbf{Alistair Plum}$^2$, \textbf{Achim Rettinger}$^1$ \\
  $^1$Trier University, Trier, Germany \\
  $^2$University of Luxembourg, Esch-sur-Alzette, Luxembourg \\
  \texttt{\{schwager, muenker, rettinger\}@uni-trier.de} \\
  \texttt{alistair.plum@uni.lu}
}

\begin{document}
\maketitle
\begin{abstract}
The transition of Large Language Models (LLMs) from exploratory tools to active ``silicon subjects'' in social science lacks extensive validation of operational validity. This study introduces Conditioned Comment Prediction (CCP), a task in which a model predicts how a user would comment on a given stimulus by comparing generated outputs with authentic digital traces. This framework enables a rigorous evaluation of current LLM capabilities with respect to the simulation of social media user behavior. We evaluated open-weight 8B models (Llama3.1, Qwen3, Ministral) in English, German, and Luxembourgish language scenarios. By systematically comparing prompting strategies (explicit vs. implicit) and the impact of Supervised Fine-Tuning (SFT), we identify a critical form vs.~content decoupling in low-resource settings: while SFT aligns the surface structure of the text output (length and syntax), it degrades semantic grounding. Furthermore, we demonstrate that explicit conditioning (generated biographies) becomes redundant under fine-tuning, as models successfully perform latent inference directly from behavioral histories. Our findings challenge current ``naive prompting'' paradigms and offer operational guidelines prioritizing authentic behavioral traces over descriptive personas for high-fidelity simulation.
\end{abstract}

\input{content/00-introduction}
\input{content/01-background}
\input{content/02-methods}
\input{content/03-experiment}
\input{content/04-conclusion}
\input{content/10-postscript}

\bibliography{custom}
\appendix
\input{content/11-appendix}

\end{document}

%% file: content/00-introduction.tex
\section{Introduction}
\label{sec:introduction}
The deployment of Large Language Models (LLMs) in computational social science is shifting from exploratory analysis to active modeling. Researchers are increasingly aiming to use these models as ``silicon subjects'' to replicate survey demographics \cite{wang2025evaluating} or model discourse dynamics \cite{zhang2025spark}. The validity of such applications rests on a fundamental assumption: that instruction-tuned models can accurately predict how specific individuals would respond to (new) stimuli.

However, the methodology for this conditioning remains largely heuristic. The dominant practice, which we refer to as \emph{explicit conditioning}, relies on describing a user's attributes in the prompt to the model (e.g.``You are a conservative voter''). This approach assumes that a model's interpretation of these labels aligns with the complex response patterns of actual individuals. This assumption is rarely tested against a ground truth. While such methods often achieve surface plausibility by generating text that looks like a social media comment, they lack operational validity: the demonstrated ability to reproduce the specific patterns of the authentic user \cite{larooij2025validation}.

In this work, we address this gap by benchmarking \textbf{Conditioned Comment Prediction} (CCP), which we view as a foundational proxy task for broader social media user simulation. Instead of attempting a full-scale simulation of user agency, we isolate the specific capability of response generation: \textit{Can the model accurately predict a user's reply to a given stimulus, based solely on the provided conditioning context?}

We systematically evaluate open-weight LLMs (8B parameter class) in three languages and their cultural environments: English, German, and Luxembourgish. By comparing prompting strategies and assessing the impact of Supervised Fine-Tuning (SFT) across lexical (ROUGE, BLEU) and semantic metrics (Embedding Distance), we aim to determine the limits of current model capabilities and the factors that drive alignment.

\subsection{Research Questions} Our investigation is guided by two primary research questions:
\begin{description}
    \setlength{\itemsep}{0pt}
    \setlength{\parskip}{0pt}
    \setlength{\parsep}{0pt}
    \item[$RQ_1$] How effectively can instruction-tuned LLMs predict authentic user comments across varying linguistic resource tiers?
    \item[$RQ_2$] Does Supervised Fine-Tuning (SFT) universally improve prediction fidelity, or is its effectiveness constrained by the models' capabilities in the target language?
\end{description}
    
\subsection{Contributions} Our work makes the following contributions to the evaluation of LLM-based user modeling:

\paragraph{Multilingual Benchmarking of Comment Prediction} We present an extensive evaluation of response generation on authentic digital traces. Unlike prior studies that focused primarily on English, our inclusion of German and Luxembourgish reveals that predictive performance is sensitive to the models' language capabilities. We identify a form-content decoupling in low-resource settings, where models fine-tuned on user data mimic the statistical texture of speech without grounding it in the user's semantic intent.

\paragraph{Evaluating Conditioning Strategies} We systematically compare the performance of explicit conditioning (conditioning on descriptions) against implicit conditioning (conditioning on behavioral history). Our results challenge the utility of biography-based approaches, showing that conditioning models directly with behavioral examples consistently yields higher fidelity. This suggests that allowing the model to perform ``latent inference'' from history is a more robust mechanism than relying on natural language descriptions.

\paragraph{Operational Guidelines} Based on our benchmarking results, we derive concrete guidelines for computational social scientists. We outline where off-the-shelf prompting suffices versus where it actively misleads, providing a roadmap for more valid and reproducible research designs.

%% file: content/01-background.tex
\section{Background}
\label{sec:background}

\subsection{LLMs as Agents in Social Simulations}
Social simulation has long been constrained by the trade-off between behavioral realism and computational tractability. Traditional agent-based models rely on hand-crafted rules that capture aggregate patterns but struggle to reproduce the nuanced, context-dependent behavior of real individuals \cite{macal2009agent}. LLMs offer a potential solution: models pre-trained on massive corpora of human text possess implicit representations of linguistic style \cite{durandard2025llms}, rhetorical strategies \cite{khan2024debating}, and even ideological positioning \cite{rottger2024political}. Recent work has demonstrated that these capabilities can be harnessed for social simulation tasks ranging from modeling network dynamics to simulating online discourse \cite{andreas2022language, hu2025simbench}.

However, the field faces a validation crisis. Despite the growing adoption of LLM-based agents in social science applications, suitable methods to assess simulation fidelity remain limited. Many studies rely on surface-level validation techniques, human raters judging ``plausibility'' or aggregate statistical properties, that fail to capture whether models genuinely reproduce individual-level behavioral patterns \cite{larooij2025validation}. The opacity of LLMs, their stochastic generation process, and documented cultural biases compound these concerns.

Our work addresses this validation gap by grounding the evaluation with respect to its operational validity: we measure alignment against actual user behavior rather than abstract notions of plausibility. By framing response generation as a prediction task, we evaluate whether a model can anticipate how a specific individual would respond to a given stimulus.

\subsection{Prompting Social Media Users}
A central challenge in persona-based simulation is determining how user characteristics should be represented and provided to the model. The literature presents two paradigms:

\paragraph{Explicit} (biography-based approaches) that operationalize personas as natural language descriptions of user attributes \cite{yu2024affordable, liu2024skepticism}. This approach draws inspiration from traditional survey-based modeling in social science. Practitioners construct \cite{liu2024skepticism} or infer \cite{gao2023s3} textual profiles specifying demographic characteristics, ideological positions, communication styles, and behavioral patterns. The model is then instructed to ``role-play'' this persona through appropriate system prompts.

\paragraph{Implicit} (history-based approaches) conditions models directly on behavioral traces, actual examples of the user's prior actions, without explicit characterization \cite{munker2025don}. This paradigm aligns with behavioral economics, which emphasizes revealed preferences over stated attributes. Rather than telling the model ``this user is politically conservative'', implicit profiling provides examples: ``this user wrote X in response to Y''. The model must perform latent inference, extracting the underlying behavioral signature from demonstrated patterns.

\paragraph{} The empirical question of which approach yields a higher fidelity simulation and under what conditions remains largely unexplored. Our work directly addresses this gap through the controlled comparison of explicit, implicit, and combined conditioning strategies.

%% file: content/02-methods.tex
\section{Methods for Conditioned Comment Prediction}
\label{sec:methods}

\subsection{Task Definition}
\label{sec:methods:task}
The CCP task is about predicting how a specific user would respond to a given stimulus (a post or a news article; see Table~\ref{tab:demo_cases} for examples). By comparing predicted responses with authentic ones, we assess whether models can capture individual-level behavioral patterns rather than producing generic responses. This framing follows the operational validity criterion: alignment should be measured against the actual individuals being simulated, not abstract notions of plausibility \cite{larooij2025validation}.

\subsection{Conditioning Strategies}
We evaluate three conditioning strategies, varying whether user characteristics are provided explicitly (via profile descriptions), implicitly (via behavioral examples), or both. This allows us to disentangle the model's ability to follow instructions about a persona from its ability to infer one.

\paragraph{User History (Implicit)} We provide up to 30 stimulus–response pairs from the original user, formatted as previous prompt-completion turns in the LLM’s native chat structure. The model receives no explicit description of the user, only examples of how they responded previously. This tests implicit conditioning: whether models can infer and reproduce user characteristics from behavioral patterns alone, without explicit instruction.

\paragraph{Generated Biography (Explicit)} We prompt \texttt{Qwen3-235B-A22B-Instruct-2507} \cite{qwen3technicalreport} to infer a short profile from up to 30 authentic comments (Appendix~\ref{sec:appendix:prompts:profiler}). The profile covers four dimensions: (1) \textit{Basics}, demographic indicators, and account type; (2) \textit{Language}, linguistic repertoire, formality, and stylistic markers; (3) \textit{Worldview}, ideologies, and group alignments; (4) \textit{Behavior}, engagement patterns, argumentation style, and communication goals. This tests explicit conditioning: whether natural-language persona descriptions suffice for faithful simulation. It also serves as a proxy for what we call ``naive prompting'', conditioning on stated attributes, without proper alignment or evaluation.

\paragraph{Combined} We provide both the inferred profile and the behavioral history. This tests whether explicit and implicit signals are complementary (yielding additive gains), redundant (history subsumes what the biography provides), or interfering (conflicting signals degrade performance).

\paragraph{Control} We provide neither behavioral history nor a generated profile, conditioning the model solely on the incoming stimulus and a generic system instruction. This serves as a baseline to isolate the impact of personalization, verifying whether improved metrics stem from actual user alignment or simply the model's general capability to generate plausible social media content.

\subsection{Models and Fine-Tuning}

\paragraph{Base Models} We evaluate three instruction-tuned models: \texttt{Llama-3.1-8B-Instruct} \cite{grattafiori2024llama}, \texttt{Qwen3-8B} without reasoning \cite{qwen3technicalreport}, and \texttt{Ministral-8B-Instruct-2410}. All models are comparable in parameter count, but differ in architecture, training data, and alignment procedures. These serve as baselines representing standard prompted persona simulation\footnote{For the remainder of this paper, we refer to these models simply as \texttt{Llama3.1}, \texttt{Qwen3}, and \texttt{Ministral}, omitting specific version suffixes for brevity.}.

\paragraph{Fine-Tuning} We apply Supervised Fine-Tuning (SFT) to all three base models on the task described in Section~\ref{sec:methods:task}. To ensure comparability across models, we use identical hyperparameters: one epoch, a maximum sequence length of 4,500 tokens, and training on complete input sequences (system prompt, user prompts, and model completions). We use the paged AdamW optimizer with 8-bit quantization \cite{dettmers20218} to enable training on a single NVIDIA L40S GPU (48GB VRAM). All remaining hyperparameters follow the TRL defaults \cite{vonwerra2022trl}.

\subsection{Datasets}

\paragraph{German ($\mathbb{X}$)}  
We use German $\mathbb{X}$ data collected around keywords related to German political discourse during the first half of 2023. The raw corpus contains 3.38M tweets comprising original posts and first-order replies from users engaging with political content.

\paragraph{English ($\mathbb{X}$)}  
The English corpus comprises 7.79M tweets, collected from $\mathbb{X}$ up to August 2023. Users were sampled by identifying 100 politically active accounts (those recently replying to U.S. politicians' content) and merging their complete followee networks, extracting up to 3,200 tweets and replies per user.

\paragraph{Luxembourgish (RTL Comments)}  
The corpus of Luxembourgish text comprises 1.02M user comments, posted by 21,427 users. The comments are published on the website of RTL\footnote{\url{https://rtl.lu}}, the main news broadcaster of Luxembourg, and were posted in the period 2012 to 2024. The topics are closely related to the corresponding news articles. Platform administrators moderate the comments; therefore, harmful, abusive, offensive, etc. content is not included.

\paragraph{Pre-Processing}
We apply uniform preprocessing across all three corpora. First, we retain only first-order replies and group them with their parent stimuli (tweets or articles), then reorganize samples by user to enable user-level modeling. We model the users strictly as repliers; the stimuli are posts by others or articles. We remove stimulus–response pairs containing URLs, images, or GIFs as these cannot be processed by text-only models. To standardize conditioning across users, we impose a maximum history size of 30 stimulus–response pairs. For users with more than 30 available interactions, we retain only the last 30 and discard the remainder. Users with fewer than four interactions are excluded, as models cannot reliably infer behavioral patterns from extremely sparse histories.

\paragraph{Splits \& Size}
We partition data at the user level so that all stimulus–response pairs from a single user appear exclusively in training or testing. This prevents cross-user leakage and enables the evaluation of cross-user generalization. From each language-specific corpus we sample $3,800$ users for training and $650$ users for testing. All sampling is deterministic, using a fixed random seed to ensure exact replication.

\paragraph{Generation \& Evaluation}
For evaluation, we always predict the last response from the user in the history retained. During both prompting and fine-tuning, the model receives the preceding retained stimulus–response pairs as chat-style prompt–completion turns (minimum 3; maximum 29). The biography (when used) is inferred from the same retained history but explicitly excludes the held-out target reply to avoid information leakage. For each model, we generate five test runs using a uniform decoding temperature of $0.75$ and $500$ max new tokens.

\subsection{Metrics}
We evaluate model performance by comparing generated replies to the corresponding authentic user responses across five independent generation runs per model. For each run, the model produces one completion for every test instance, and we compute all metrics over the full set of authentic–generated reply pairs. We then aggregate results across runs, reporting the mean and standard deviation for every metric–model combination. This procedure captures both the overall performance and the stochastic variability introduced by sampling-based generation. Extended results, including standard deviations and evaluations with alternative embedding models, are reported in the Appendix~\ref{sec:app:ext}.

\paragraph{Embedding Distance}
To assess semantic alignment between generated and authentic user replies, we compute the cosine distance between their embedding representations. Our primary embedding model is \texttt{Qwen3-Embedding-8B} \cite{qwen3embedding}. We averaged the scores over the whole run. This metric captures similarity in communicative intent and discourse structure. Distances range from 0 to 2, with lower values indicating closer approximation of the target user’s response profile. 

\paragraph{ROUGE-1}
We compute ROUGE-1 (unigram overlap) \cite{lin2004rouge} to quantify the lexical similarity between the generated and authentic responses. This surface-level metric reflects the model’s ability to reproduce user-specific lexical choices, including vocabulary, named entities, and hashtag usage. 

\input{figures/demo}

\paragraph{BLEU}
We report BLEU \cite{papineni2002bleu} to measure the precision-oriented n-gram overlap between generated and authentic replies. BLEU captures the model’s ability to reproduce user-specific multiword expressions and stable phrasing patterns. 

\paragraph{Length Ratio (LR)}
We report the length ratio as derived from the standard BLEU calculation \cite{papineni2002bleu}. This metric is calculated as the ratio of the length generated by the system to the reference length ($ratio = \frac{len_{gen}}{len_{ref}}$). It quantifies the difference in output volume between the model and the authentic user, where a value of $1.0$ indicates perfect alignment in length regardless of content overlap.

%% file: figures/demo.tex
\begin{table*}[t]
    \centering
    \renewcommand{\arraystretch}{1.4}
    \setlength{\tabcolsep}{3pt}
    \begin{tabular}{p{4.6cm} p{3cm} p{3cm} c p{3cm} c}
    \toprule
    & & \multicolumn{2}{c}{\textbf{Base Model}} & \multicolumn{2}{c}{\textbf{Fine-Tuned Model}} \\
    \cmidrule(lr){3-4} \cmidrule(lr){5-6}
    \textbf{Stimulus} & \textbf{Authentic Reply} & \textbf{Reply} & \textbf{D} & \textbf{Reply} & \textbf{D}\\
    \midrule
    \raggedright \textgreater{}@User1: .@User2 is trying to turn your kids into BLM \& LGBTQ+ activists... features a drag queen. Skittles have gone completely woke. & 
    \raggedright @User1 Never really liked Skittles. Now I know why. Pathetic & 
    \raggedright @User1 What a f****** joke. I bet you are a total loser in life. & 
    .27 & 
    \raggedright @User1 @User2 Now I know why I never liked them & 
    .08 \\
    \midrule
    \raggedright \textgreater{}@User1: NEWS \textit{[siren]}: It's official, NASA says July was the hottest month ever recorded on Earth & 
    \raggedright @User1 LOL & 
    \raggedright @User1 By a landslide in the land of make believe & 
    .26 & 
    \raggedright @User1 LOL the Moon??? & 
    .13 \\
    \midrule
    \raggedright \textgreater{}@User1: I just left my parents house where... my father passed away. I am going to work today because I'm not sure what else to do... & 
    \raggedright @User1 I'm so sorry for your loss, \textit{[NAME]}. & 
    \raggedright @User1 Sorry to hear that about your dad. \textit{[broken heart]} Stay strong... & 
    .29 & 
    \raggedright @User1 So sorry for your loss. & 
    .16 \\
    \midrule
    \raggedright \textgreater{}@User1: The timeline does not lie. @User2 has slow-walked this country to the brink of default... & 
    \raggedright @User1 @User2 You are in way over your head. Enjoy this fleeting moment of power. & 
    \raggedright @User1 @User2 He's a puppet. & 
    .39 & 
    \raggedright @User1 @User2 What does this have to do with anything? & 
    .32 \\
    \midrule
    \raggedright \textgreater{}@User1: Oh great, another meeting that could have been an email. & 
    \raggedright @User1 \textit{[rofl]} Story of my life. & 
    \raggedright @User1 That is annoying. & 
    .22 & 
    \raggedright @User1 You should be grateful you have a job. & 
    .58 \\
    \bottomrule
    \end{tabular}
    \caption{\textbf{Qualitative comparison of selected reply predictions.} The table presents the input Stimulus, the Authentic Reply, and generated responses from the Base and Fine-Tuned versions of \textbf{\texttt{Llama-3.1-8B}}. Columns labeled \textbf{D} denote the embedding distance to the authentic reply (lower is better), calculated using \texttt{Qwen3-Embedding-8B} \cite{qwen3embedding}. All samples are in \textbf{English} using the \textit{Biography+History} conditioning strategy; note that the behavioral histories used for conditioning are omitted from this display for brevity.  Usernames are anonymized and emojis are replaced with descriptions like \textit{[party]}.}
    \label{tab:demo_cases}
\end{table*}

%% file: content/03-experiment.tex
\section{Experiments}
\label{sec:experiments}
\input{tables/overview-table}
This section presents the results of our CCP experiments by organizing the discussions along our main research questions.
We report performance metrics for lexical overlap (BLEU, ROUGE-1), semantic alignment (embedding distance) and generation constraints (length ratio). All results represent the mean over five independent runs. 

\subsection{Prediction Fidelity ($RQ_1$ \& $RQ_2$)}
\label{sec:exp:fidelity}

Table~\ref{tab:rq1_rq2} summarizes the performance of base and fine-tuned (FT) models in English (EN), German (DE), and Luxembourgish (LB).

\paragraph{Baseline Capabilities and Language Hierarchy} Addressing $RQ_1$, we observe a strict performance hierarchy dictated by linguistic resource tiers. In English, base models exhibit non-trivial alignment (BLEU $0.053$, embedding distance $0.420$), indicating a grounding for both the syntax and semantics of the domain. This capability degrades moderately for German and strongly for Luxembourgish (BLEU $\approx 0.003$). Crucially, the low absolute values across all metrics underscore the inherent difficulty of the task: predicting exact social media replies is a high-entropy challenge constrained by partial observability. Models must not only capture individual variance, but also contend with significant uncertainty arising from unobserved external stimuli that drive actual behavior.

\paragraph{The Effectiveness of Fine-Tuning}
For the dominant language (EN), supervised fine-tuning acts as a capability amplifier. \texttt{Llama3.1} achieves substantial gains in lexical alignment (BLEU $0.053 \rightarrow 0.083$) while simultaneously tightening semantic alignment (embedding distance $0.420 \rightarrow 0.397$), as illustrated qualitatively in Table~\ref{tab:demo_cases}. However, this effect is less consistent in German. While lexical metrics improve (BLEU $0.065 \rightarrow 0.095$), the semantic alignment remains stagnant (embedding distance $\approx 0.50$), suggesting that SFT refines style but struggles to deepen semantic grounding beyond the base model's capabilities.

\input{tables/component-comparison-table}

\paragraph{Form-Content Decoupling in Low-Resource Settings}
A critical divergence appears in Luxembourgish. Although SFT significantly improves surface-level metrics (BLEU and ROUGE-1), it degrades semantic alignment (the embedding distance increases from $0.579 \rightarrow 0.605$ for \texttt{Llama3.1}). We interpret this as a decoupling of form and content due to a lack of underlying robustness in the pre-trained representation. The base models produce erratic output lengths (length ratio $\approx 2.98$ for \texttt{Ministral}); SFT successfully constrains the model to the correct length distribution (length ratio $\approx 1.07$) and improves the n-gram statistics, but the increasing embedding distance suggests that the model is simply mimicking the structure of the language rather than retaining semantic fidelity. Critically, this observation is also consistent with the embeddings generated by \texttt{LuxEmbedder} \cite{philippy-etal-2025-luxembedder} (see Appendix~\ref{sec:app:ext}), confirming that the semantic degradation is due to the fine-tuning process rather than an artifact of a specific evaluation metric.

\paragraph{Model Comparison}
\texttt{Llama3.1} demonstrates superior stability across all languages. Crucially, it is the only base model that maintains a realistic length ratio ($1.11$ in EN, $1.29$ in LB), whereas \texttt{Qwen3} and \texttt{Ministral} suffer from severe verbosity (e.g., \texttt{Ministral} LB length ratio $2.98$), generating text that is structurally completely misaligned with the target domain. While \texttt{Ministral} shows the highest alignment scores in Luxembourgish after fine-tuning, its inability to adhere to length constraints without fine-tuning makes it practically unusable for simulation.

\subsection{Ablation Study: Implicit vs. Explicit Conditioning}
\label{sec:exp:conditioning}
\input{figures/history-length-impact}

Table~\ref{tab:rq3} isolates the impact of conditioning strategies (Control, User History, Generated Biography, and Combined) using \texttt{Llama3.1} in the English dataset.

\paragraph{Zero-Context Baseline Evaluation} The Control condition establishes the lower performance limit, representing a model that replies to the stimulus without any user-specific context. Interestingly, fine-tuning on the Control condition alone yields a competitive ROUGE-1 score ($0.207$), suggesting that a significant portion of lexical predictability is driven solely by the topic of the stimulus and general adaptation to the style of user comments. However, the semantic alignment remains weaker (embedding distance $0.418$) compared to user-conditioned models ($0.399$ for History). This indicates that while the model can learn the general ``shape'' of a reply, it requires user-specific conditioning to accurately capture the writing style, specific stance and semantic intent of the individual.

\paragraph{Structural Misalignment in Explicitly Conditioned Base Models} With the base model, the Biography-Only strategy fails catastrophically, exhibiting a length ratio of $4.907$. This failure stems from a lack of structural grounding: without the few-shot examples provided by the history, the model fails to infer the structural constraints of the platform (e.g., brevity, informality). It generates content relevant to the persona but fails to adopt the format of a social media reply. Fine-tuning corrects this (LR $\rightarrow 0.935$), indicating that SFT is crucial to teach models how to map explicit persona descriptions into the correct output format.

\paragraph{Latent Inference via Fine-Tuning}
The most significant finding is the redundancy of explicit conditioning in the fine-tuned setting. Although the Biography-Only condition performs poorly with the base model, the History-Only condition is relatively robust. After fine-tuning, the performance gap between History-Only (emb. dist. $0.399$) and Biography+History (emb. dist. $0.397$) is marginal. This suggests that SFT enables the model to perform latent inference: extracting latent behavioral vectors directly from the history. The model learns to infer the persona from behavioral traces just as effectively as it utilizes a pre-generated biography. Consequently, for fine-tuned models, the computational cost of profiling in an additional step yields diminishing returns compared to simply conditioning on raw history.

\subsection{Ablation Study: Sensitivity to History Length}

Figure~\ref{fig:rq4} illustrates the trajectory of model performance as the number of behavioral examples available increases from 0 to 29. We evaluate this using the History-Only condition to isolate the impact of behavioral context scaling.

\paragraph{Solving the ``Cold Start'' Problem} The most immediate distinction between the base and the fine-tuned models appears in the low-context regime ($N < 5$). The base model exhibits extreme volatility without context: at $N=0$, the length ratio spikes above $4.4$ and embedding distance degrades above $0.6$, indicating that the model fails to adhere to the platform's constraints. It relies entirely on In-Context Learning (ICL) to infer the format, requiring approximately 5 examples to stabilize. In contrast, the fine-tuned model shows zero-shot stability. Even with no history ($N=0$), it maintains a good length ratio ($\approx 1.1$) and a superior semantic alignment. This confirms that SFT effectively internalizes the platform's structural priors and the general semantic distribution of the user base, decoupling basic simulation competence from the availability of history.

\paragraph{Scaling and Non-Saturation} Contrary to expectations of diminishing returns, we do not observe a distinct saturation point for our metrics. BLEU and ROUGE-1 scores for the FT model exhibit an upward trend throughout the 29-turn window. This suggests that user behavior in this domain is sufficiently complex that a window of 29 interactions does not exhaust the predictive signal; each additional historical data point continues to refine the simulation. The apparent volatility and performance drop in the extreme tail ($N = 28$) coincides with a decrease in sample size (represented by the background histogram), which could render those specific fluctuations statistical artifacts rather than the true performance degradation.

%% file: tables/overview-table.tex
\begin{table*}
\centering
\begin{tabular}{@{}llcccccccc@{}}
\toprule
 & & \multicolumn{2}{c}{\textbf{BLEU} ($\uparrow$)} & \multicolumn{2}{c}{\textbf{Len. Ratio} ($\to 1$)} & \multicolumn{2}{c}{\textbf{ROUGE-1} ($\uparrow$)} & \multicolumn{2}{c}{\textbf{Emb. Dist.} ($\downarrow$)} \\
\cmidrule(lr){3-4} \cmidrule(lr){5-6} \cmidrule(lr){7-8} \cmidrule(lr){9-10}
\textbf{Lang} & \textbf{Model} & \textbf{Base} & \textbf{FT} & \textbf{Base} & \textbf{FT} & \textbf{Base} & \textbf{FT} & \textbf{Base} & \textbf{FT} \\
\midrule
\multirow{3}{*}{EN} 
 & Llama3.1 & 0.053 & \textbf{0.083} & 1.110 & 0.961 & 0.190 & \textbf{0.229} & 0.420 & \textbf{0.397} \\
 & Qwen3 & 0.038 & 0.081 & 1.624 & 0.933 & 0.180 & 0.220 & 0.418 & 0.408 \\
 & Ministral & 0.039 & 0.081 & 1.428 & \textbf{0.985} & 0.186 & 0.223 & 0.424 & 0.404 \\
\midrule
\multirow{3}{*}{DE} 
 & Llama3.1 & 0.065 & \textbf{0.095} & 1.205 & 0.915 & 0.172 & \textbf{0.192} & 0.509 & 0.504 \\
 & Qwen3 & 0.049 & 0.094 & 1.633 & 0.926 & 0.171 & 0.188 & 0.509 & 0.512 \\
 & Ministral & 0.046 & 0.087 & 1.627 & \textbf{1.073} & 0.160 & 0.182 & 0.505 & \textbf{0.502} \\
\midrule
\multirow{3}{*}{LB} 
 & Llama3.1 & 0.007 & 0.009 & 1.291 & 0.897 & 0.113 & 0.108 & 0.579 & 0.605 \\
 & Qwen3 & 0.003 & 0.008 & 2.427 & 0.886 & 0.079 & 0.107 & \textbf{0.578} & 0.610 \\
 & Ministral & 0.003 & \textbf{0.010} & 2.980 & \textbf{1.077} & 0.081 & \textbf{0.114} & 0.583 & 0.597 \\
\bottomrule
\end{tabular}
\caption{\textbf{Multilingual Performance Evaluation (RQ1 \& RQ2).} Results show the impact of Supervised Fine-Tuning (FT) vs. prompting the base model (Base) on prediction quality. Best values per comparison unit are \textbf{bolded}. Reported values are the mean across 5 independent generation runs on a hold-out test set of 650 users. All models (8B parameters) were conditioned using the combined \textit{Biography+History} strategy and trained on a dataset of 3,800 users per language. Extended results including standard deviations and other embedding models in Appendix~\ref{sec:app:ext}.}
\label{tab:rq1_rq2}
\end{table*}

%% file: tables/component-comparison-table.tex
\begin{table*}
\centering
\begin{tabular}{@{}lcccccccc@{}}
\toprule
 & \multicolumn{2}{c}{\textbf{BLEU} ($\uparrow$)} & \multicolumn{2}{c}{\textbf{Len. Ratio} ($\to 1$)} & \multicolumn{2}{c}{\textbf{ROUGE-1} ($\uparrow$)} & \multicolumn{2}{c}{\textbf{Emb. Dist.} ($\downarrow$)} \\
\cmidrule(lr){2-3} \cmidrule(lr){4-5} \cmidrule(lr){6-7} \cmidrule(lr){8-9}
\textbf{Conditioning} & \textbf{Base} & \textbf{FT} & \textbf{Base} & \textbf{FT} & \textbf{Base} & \textbf{FT} & \textbf{Base} & \textbf{FT} \\
\midrule
Control & 0.004 & 0.076 & 4.418 & \textbf{1.000} & 0.079 & 0.207 & 0.615 & 0.418 \\
Bio & 0.005 & 0.079 & 4.907 & 0.935 & 0.084 & 0.220 & 0.513 & 0.407 \\
History & 0.054 & 0.077 & 1.118 & 1.094 & 0.182 & \textbf{0.229} & 0.428 & 0.399 \\
Bio + History & 0.053 & \textbf{0.083} & 1.110 & 0.961 & 0.190 & 0.229 & 0.420 & \textbf{0.397} \\
\bottomrule
\end{tabular}
\caption{\textbf{Impact of conditioning strategies.} Results compare the performance of explicit conditioning (Biography) versus implicit conditioning (History) for \textbf{\texttt{Llama-3.1-8B}} in \textbf{English}. Best values are \textbf{bolded}. Reported values are the mean across 5 independent generation runs on a hold-out test set of 650 users. All models were trained on a dataset of 3,800 users.}
\label{tab:rq3}
\end{table*}

%% file: figures/history-length-impact.tex
\begin{figure*}[ht]
    \centering
    \includegraphics[width=1.0\linewidth]{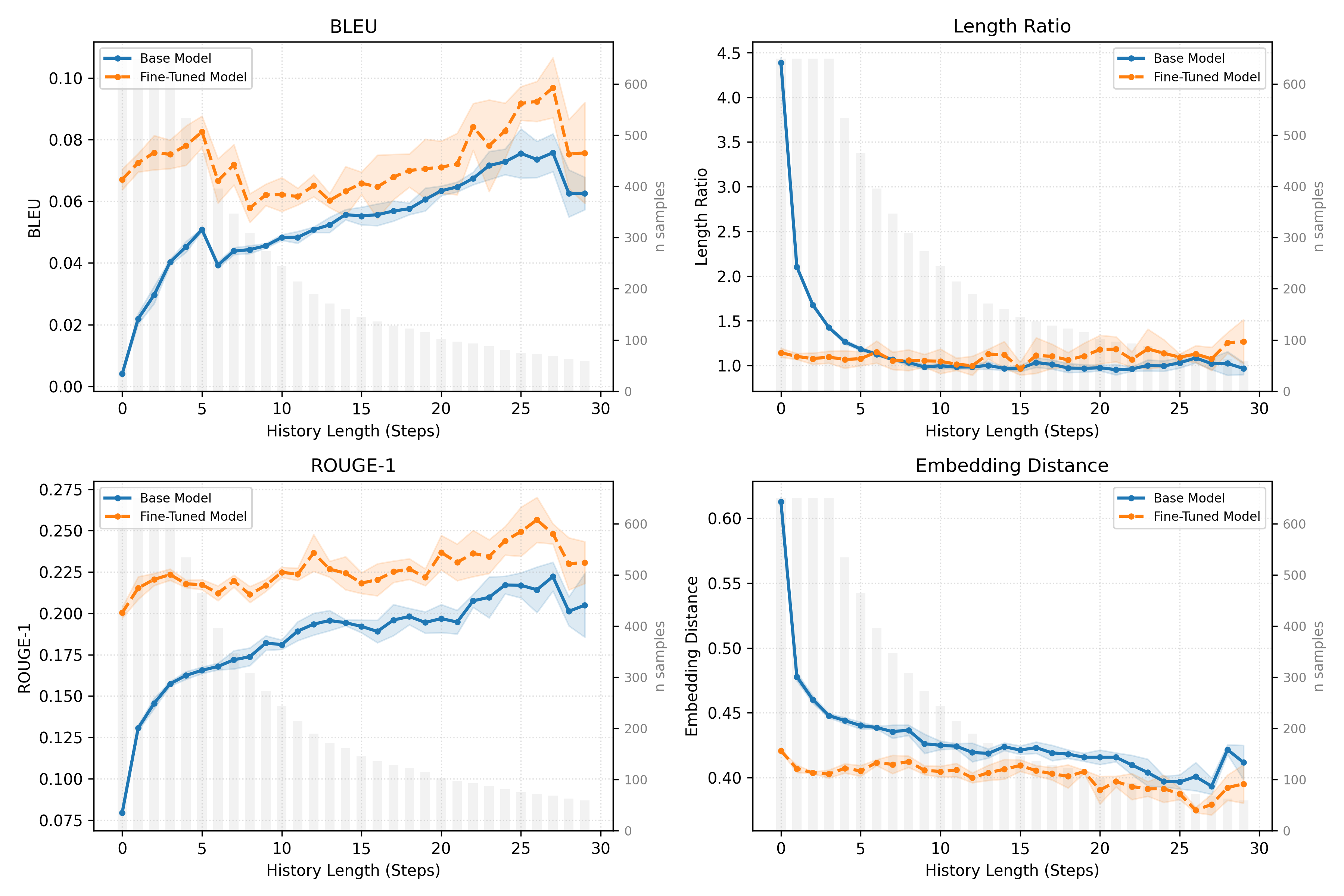}
    
    \caption{\textbf{Impact of history length on predictive performance.} Results illustrate the dependence between the volume of provided behavioral history (number of previous comments) and prediction quality. Shaded regions represent the standard deviation across 5 independent generation runs, while the underlying gray bars indicate the sample size distribution per length bucket. Analysis is based on \textbf{\texttt{Llama-3.1-8B}} in \textbf{English} using the \textit{History-Only} conditioning strategy. The fine-tuned model was trained on a dataset of 3,800 users.}
    \label{fig:rq4}
\end{figure*}

%% file: content/04-conclusion.tex
\section{Recommendations and Future Work}
\label{sec:conclusion}
In this work, we systematically evaluated the capabilities of instruction-tuned Large Language Models to perform (conditioned) comment prediction, which we consider as a sub-task on the path to accurate simulation of social media users.

\subsection{Recommendations}
\paragraph{Anchoring Model Performance via Behavioral Context} We strongly advise against using base models with explicit conditioning alone (Biography-Only), as this strategy consistently leads to structural failure and extreme verbosity (LR $\approx 4.9$). If authentic digital traces are not available, practitioners should provide generic behavioral demonstrations (general history). Even non-specific examples could serve as critical ``structural anchors'', enabling the model to adapt to the domain's format and length constraints, thereby stabilizing performance.

\paragraph{Prioritizing Authentic Behavioral Data} While a generic history stabilizes the structure, authentic digital traces remain the gold standard for improving simulation fidelity. Our results indicate that conditioning on actual user behavior provides a dual benefit: it enforces structural compliance (like generic history) while simultaneously maximizing semantic and lexical alignment (unlike generic history). Whenever available, raw behavioral logs should take precedence over synthetic user descriptions. Furthermore, this approach mitigates the potential for researcher bias inherent in the subjective construction of explicit personas and the intensive prompt-engineering typically required for behavioral alignment.

\paragraph{Limitations of SFT in Non-English Contexts} We caution that SFT is not a universal solution for all linguistic environments. In our experiments with 8B-parameter models, SFT proved difficult for German and Luxembourgish. Although it successfully corrected the output length, it failed to significantly improve semantic grounding (German) or actively degraded it (Luxembourgish). Practitioners working with small- or mid-sized models in these languages should view SFT primarily as a tool for formatting control, not semantic enhancement.

\paragraph{Performance Convergence Post-Fine-Tuning}
In high-resource domains (English), SFT acts as a powerful equalizer, rendering specific architectural choices and complex conditioning strategies largely redundant. Our results show that while base models exhibit vast performance disparities (e.g., \texttt{Llama3.1} vs. \texttt{Qwen3}), fine-tuning causes them to converge to a nearly identical performance ceiling (BLEU $\approx 0.08$). Similarly, the distinct advantages of specific prompting strategies (Biography vs. History) disappear after fine-tuning. Consequently, for English applications, practitioners should prioritize data quantity and quality over model selection or prompt engineering, as SFT robustly aligns even simpler setups to the upper performance limit.

\subsection{Future Work}
\label{sec:future_work}

\paragraph{Robustness and Generalization} 
To determine the limits of our findings, future work should test the stability and requirements of user simulation. We propose expanding benchmarks to measure multi-turn stability, verifying whether persona consistency holds over prolonged interactions or succumbs to drift. Additionally, a precise quantification of the information density in the prompt required to guarantee convergence is necessary to establish the minimum data thresholds for valid simulation. Finally, the scope of evaluation must broaden to include non-verbal actions (such as liking) and richer environmental inputs, testing whether the simulation capabilities we observed can generalize to complex, multi-modal platform dynamics.

\paragraph{Scaling Laws and Model Size} 
Our observation of the form-content decoupling in Luxembourgish raises critical questions regarding model capacity. It remains unclear whether the failure to ground semantics is an inherent limitation of SFT in low-resource settings or an artifact of the 8B parameter scale. Small models are known to have fragile weight constellations. Future work must investigate whether larger models ($\geq$ 70B), which presumably possess more robust representations for German and Luxembourgish, can overcome this decoupling of form and content.

\paragraph{Semantic Alignment in Training} 
The observed divergence between lexical overlap and semantic grounding, which is most acute in our low-resource experiments, suggests that standard cross-entropy loss is insufficient for user simulation in uncertain or sparse data scenarios. Current training paradigms encourage models to minimize perplexity (surface-level mimicry) rather than maximizing semantic fidelity. Future research should develop and test training objectives that directly optimize for semantic alignment, such as Direct Preference Optimization (DPO), where the loss function explicitly penalizes semantic distance from the target user's discourse history.

%% file: content/10-postscript.tex
\section*{Limitations}
\label{sec:limitations}

\paragraph{Lacking Comparability between Languages} While we benchmark performance across three languages, we acknowledge that these tasks are not strictly comparable. The predictive signal in the input (the prompt) and the variety in the output (the completion) may vary strongly between the different dataset types. Consequently weaker prediction fidelity in German and Luxembourgish may reflect higher unpredictability of that specific dataset rather than purely linguistic deficiencies in the models.

\paragraph{Reliance on Automated Metrics}
Our evaluation relies exclusively on automated metrics (BLEU, ROUGE, Embedding Distance). While embedding distance serves as a robust proxy for semantic grounding, it cannot fully capture nuanced persona failures, such as tonal drift or subtle hallucinations, that a human would identify. 

\paragraph{Profiler Dependency}
The Generated Biography condition utilizes a profiler to create explicit biographies. We acknowledge this represents a form of ``naive prompting'' which may not be informationally optimal compared to highly curated expert prompts. However, the performance gains observed after Supervised Fine-Tuning confirm that these generated bios do encode the relevant signal, even if base models struggle to utilize it zero-shot. We therefore treat this condition as a representative baseline for standard automated profiling, noting that an exhaustive evaluation of prompt engineering strategies, as well as comparisons against socio-demographic profiles utilizing data beyond strictly inferable attributes, remain beyond the scope of this study.

\paragraph{Model Selection and Scale}
We deliberately restricted our evaluation to the 8B-parameter class of open-weight models to ensure reproducibility and align with the resource constraints. However, this focus imposes a constraint on model capacity. As observed in our Luxembourgish results, the decoupling of structural form and semantic content may be limited to this specific scale. Our findings, therefore, may not fully extrapolate to frontier-scale proprietary models.

\section*{Ethics}
While our work aims to advance scientific understanding of LLM behavior and establish methodological standards for social simulation, we acknowledge that the techniques we systematically optimize can be repurposed for harmful ends.

\subsection*{Dual Use: Fake News/Misinformation}
The most immediate concern is that improved user simulation enables more sophisticated forms of online manipulation. Our work demonstrates that LLMs can generate content that mimics individual communication patterns with measurable fidelity. Malicious actors could exploit these capabilities for:

\paragraph{Coordinated Inauthentic Behavior} Generating large volumes of synthetic social media content that appears to originate from diverse, authentic users. Unlike traditional bot campaigns that rely on template-based generation or simple text spinning, LLM-based simulation can produce varied, contextually appropriate responses that evade simple detection heuristics. Our finding that fine-tuned models achieve strong performance even with limited user history (5-10 examples) is particularly concerning because adversaries need not compromise entire accounts but merely scrape public posting histories to create convincing impersonations.

\paragraph{Micro-Targeted Disinformation} Tailoring persuasive content to specific demographic or ideological profiles. Our profiling methodology, extracting implicit behavioral signatures from digital traces, could be inverted to craft messages designed to resonate with particular audience segments. The convergence we observe after fine-tuning means that even resource-constrained actors could deploy effective simulation systems without requiring cutting-edge models or extensive prompt engineering.

\subsection*{Privacy and Consent Considerations}
Our study utilizes real user data from $\mathbb{X}$ to train models that simulate individual responses. Although our data set consists of publicly available posts and  replies from regular users, the individuals whose data we used did not provide explicit informed consent for their communication patterns to be replicated by generative models. This raises concerns about digital privacy rights, even when dealing with public data. The simulation of specific individuals' replying behavior creates synthetic content that mimics their communication style, potentially enabling the creation of convincing but fabricated posts that could be attributed to real people.

\section*{Acknowledgments}
We thank Christoph Hau and Lotta Jaeger for constructive discussions. This study was conducted with a financial contribution from the EU’s Horizon Europe Framework (HORIZON-CL2-2022-DEMOCRACY-01-07) under grant agreement number 101095095.

%% file: content/11-appendix.tex
\section{Prompts}
\label{sec:appendix:prompts}

\subsection{User Profiler Prompt}
\label{sec:appendix:prompts:profiler}

The following system prompt is used to generate the explicit user profiles (``Bio'' condition) from the behavioral history.

\begin{tcolorbox}[title=Prompt: User Profiler, colback=white, colframe=black!75!white, fonttitle=\bfseries]
\small
You are profiling a user for LLM roleplay simulation. Another LLM will read this profile and simulate this person's responses. Write in second-person active voice: ``You are...'', ``You write...'', ``You believe...''. State what you observe directly. Do not explain your reasoning or cite evidence.

\vspace{0.5em}
\noindent\textbf{Task:} Create a 500-1000 token profile organized into four sections. Write naturally. If information is sparse, write less. If someone is unremarkable, say so.

\noindent\rule{\textwidth}{0.4pt}

\textbf{BASICS}\\
Who is this? What kind of account?\\
Demographics: age, location, education, occupation (if inferable). Account type: personal, parody, activist, professional, organizational. Authentic voice or performance?

\vspace{0.5em}
\textbf{LANGUAGE}\\
How do they use language?\\
Which languages? Code-switching patterns? Formality level? Dialect markers? Native or non-native? Distinctive style?

\vspace{0.5em}
\textbf{WORLDVIEW}\\
What do they believe?\\
Positions on issues. Ideological patterns. What they care about. Who they align with. Who they oppose. Consistency or contradiction.

\vspace{0.5em}
\textbf{BEHAVIOR}\\
How do they operate?\\
Engagement frequency and depth. Argumentation style. Tone. Who they write for. What they are trying to accomplish.

\noindent\rule{\textwidth}{0.4pt}

\textbf{Examples:}\par
\noindent\textit{[Four examples covering a traditional Baden-Württemberg professional, a vague/disengaged user, and a Luxembourgish language advocate are included here to provide diverse few-shot guidance...]}
\noindent\rule{\textwidth}{0.4pt}

Write with clarity and confidence. Make this profile useful for roleplay simulation.

\{content\}

\end{tcolorbox}

\subsection{Simulation Prompts}
\label{sec:appendix:prompts:simulation}

The \textbf{Reply Instruction} is the standardized trigger used in all experimental conditions to initiate content generation. The \textbf{System Prompt} is injected specifically for conditions without an explicit user profile (i.e., \textit{History-Only} and \textit{Control}), instructing the model to rely on its context window for behavioral consistency.

\begin{tcolorbox}[title=System Prompt (No-Bio Conditions), colback=white, colframe=black!75!white, fonttitle=\bfseries]
\small
You are a user commenting on online content. Keep your comments consistent with your previous writing style and the perspectives you have expressed earlier.
\end{tcolorbox}

\vspace{0.5em}

\begin{tcolorbox}[title=User Instruction (All Conditions), colback=white, colframe=black!75!white, fonttitle=\bfseries]
\small
Comment on the following content:

\{content\}
\end{tcolorbox}

\input{tables/multilingual-training-table}
\input{tables/model-size-table}

\section{Data, Code, and Model Availability} 
\label{sec:appendix:availability} The technical pipeline and source code is available on GitHub: \url{https://github.com/nsschw/Conditioned-Comment-Prediction}. To mitigate potential misuse while ensuring reproducibility, fine-tuned models and datasets are restricted to scientific use and shared only upon request. This policy aligns our open science commitment with responsible research practices.

\section{Additional Experiments}
\label{sec:appendix:experiments}

\subsection{Multilingual Joint Training}
\label{sec:app:multilingual}

Table~\ref{tab:rq5} contrasts the performance of models fine-tuned on a monolingual corpus (``Mono'') against one trained on a joint mixture of all three languages (``Mix'').

\paragraph{Performance Parity} The results between the Mixed and Monolingual conditions are effectively indistinguishable, with differences in BLEU and embedding distance not becoming significant. This parity suggests that the model capacity of 8B parameters is sufficient to accommodate multiple distinct linguistic distributions without suffering from interference or ``curse of multilinguality''.

\paragraph{Absence of Cross-Lingual Synergy} Crucially, however, we observe no positive transfer effects for the low-resource language. We hypothesized that joint training might allow Luxembourgish to benefit from the structural or semantic scaffolding of English and German. The lack of improvement in the Mix condition (LB BLEU $0.008$ vs. Mono $0.009$) indicates that these languages are likely being modeled in orthogonal subspaces. Although joint training is a viable strategy for the efficiency of deployment (serving one model instead of three), it does not serve as a remediation strategy for data scarcity in this domain.

\subsection{Impact of Model Size}
\label{sec:app:scaling}

Table~\ref{tab:rq6} evaluates the scaling laws of simulation fidelity using the \texttt{Qwen3} family, ranging from 0.6B to 8B parameters in the English dataset.

\paragraph{Capacity Constraints of Small Models} Small models (0.6B and 1.7B) exhibit distinct limitations. Although SFT successfully regulates their structural output, fixing the length ratio of the 1.7B Base model ($3.585 \rightarrow 1.319$), it cannot compensate for their limited semantic reasoning. Both models plateau at a BLEU score of $\approx 0.058$ and fail to significantly reduce the embedding distance ($\approx 0.420$), indicating that they are learning to mimic the format of the user's speech, but lack the capacity to capture deeper semantic patterns.

\clearpage
\onecolumn 
\section{Extended Tables} 
\label{sec:app:ext}
\input{tables/appendix_overview-table}

%% file: tables/multilingual-training-table.tex
\begin{table*}
\centering
\scriptsize
\setlength{\tabcolsep}{2.4pt}
\begin{tabular}{@{}lcccccccc@{}}
\toprule
 & \multicolumn{2}{c}{\textbf{BLEU} ($\uparrow$)} & \multicolumn{2}{c}{\textbf{Len. Ratio} ($\to 1$)} & \multicolumn{2}{c}{\textbf{ROUGE-1} ($\uparrow$)} & \multicolumn{2}{c}{\textbf{Emb. Dist.} ($\downarrow$)} \\
\cmidrule(lr){2-3} \cmidrule(lr){4-5} \cmidrule(lr){6-7} \cmidrule(lr){8-9}
\textbf{Lang} & Mix & Mono & Mix & Mono & Mix & Mono & Mix & Mono \\
\midrule
EN & 0.082 ($\pm$ 0.003) & \textbf{0.083 ($\pm$ 0.001)} & \textbf{0.964 ($\pm$ 0.067)} & 0.961 ($\pm$ 0.042) & 0.226 ($\pm$ 0.003) & \textbf{0.229 ($\pm$ 0.003)} & 0.398 ($\pm$ 0.004) & \textbf{0.397 ($\pm$ 0.001)} \\
DE & 0.094 ($\pm$ 0.001) & \textbf{0.095 ($\pm$ 0.002)} & 0.859 ($\pm$ 0.026) & \textbf{0.915 ($\pm$ 0.029)} & 0.192 ($\pm$ 0.005) & \textbf{0.192 ($\pm$ 0.003)} & \textbf{0.503 ($\pm$ 0.004)} & 0.504 ($\pm$ 0.002) \\
LB & 0.008 ($\pm$ 0.001) & \textbf{0.009 ($\pm$ 0.000)} & 0.787 ($\pm$ 0.025) & \textbf{0.897 ($\pm$ 0.030)} & \textbf{0.109 ($\pm$ 0.002)} & 0.108 ($\pm$ 0.001) & 0.606 ($\pm$ 0.002) & \textbf{0.605 ($\pm$ 0.003)} \\
\bottomrule
\end{tabular}
\caption{\textbf{Mixed vs. monolingual fine-tuning.} Results compare the performance of mixed versus monolingual fine-tuning strategies for \textbf{\texttt{Llama-3.1-8B}}. Best values are \textbf{bolded}. Reported values are Mean ($\pm$ Standard Deviation) across 5 independent generation runs on a hold-out test set of 650 users.}
\label{tab:rq5}
\end{table*}

%% file: tables/model-size-table.tex
\begin{table*}
\centering
\scriptsize
\setlength{\tabcolsep}{2.4pt}
\begin{tabular}{@{}lcccccccc@{}}
\toprule
 & \multicolumn{2}{c}{\textbf{BLEU} ($\uparrow$)} & \multicolumn{2}{c}{\textbf{Len. Ratio} ($\to 1$)} & \multicolumn{2}{c}{\textbf{ROUGE-1} ($\uparrow$)} & \multicolumn{2}{c}{\textbf{Emb. Dist.} ($\downarrow$)} \\
\cmidrule(lr){2-3} \cmidrule(lr){4-5} \cmidrule(lr){6-7} \cmidrule(lr){8-9}
\textbf{Size} & Base & FT & Base & FT & Base & FT & Base & FT \\
\midrule
0.6B & 0.027 ($\pm$ 0.000) & \textbf{0.058 ($\pm$ 0.002)} & 1.956 ($\pm$ 0.023) & \textbf{1.290 ($\pm$ 0.050)} & 0.155 ($\pm$ 0.002) & \textbf{0.206 ($\pm$ 0.002)} & 0.452 ($\pm$ 0.003) & \textbf{0.420 ($\pm$ 0.002)} \\
1.7B & 0.016 ($\pm$ 0.001) & \textbf{0.058 ($\pm$ 0.006)} & 3.585 ($\pm$ 0.150) & \textbf{1.319 ($\pm$ 0.125)} & 0.176 ($\pm$ 0.002) & \textbf{0.202 ($\pm$ 0.002)} & 0.425 ($\pm$ 0.002) & \textbf{0.421 ($\pm$ 0.002)} \\
4B & 0.041 ($\pm$ 0.002) & \textbf{0.080 ($\pm$ 0.002)} & 1.530 ($\pm$ 0.038) & \textbf{0.986 ($\pm$ 0.034)} & 0.180 ($\pm$ 0.002) & \textbf{0.216 ($\pm$ 0.002)} & 0.423 ($\pm$ 0.003) & \textbf{0.410 ($\pm$ 0.003)} \\
8B & 0.038 ($\pm$ 0.001) & \textbf{0.081 ($\pm$ 0.002)} & 1.624 ($\pm$ 0.035) & \textbf{0.933 ($\pm$ 0.057)} & 0.180 ($\pm$ 0.002) & \textbf{0.220 ($\pm$ 0.001)} & 0.418 ($\pm$ 0.001) & \textbf{0.408 ($\pm$ 0.003)} \\
\bottomrule
\end{tabular}
\caption{\textbf{Impact of model size.} Results compare performance across the \textbf{\texttt{Qwen3}} model family in \textbf{English}. Best values are \textbf{bolded}. Reported values are Mean ($\pm$ Standard Deviation) across 5 independent generation runs on a hold-out test set of 650 users.}
\label{tab:rq6}
\end{table*}

%% file: tables/appendix_overview-table.tex
\begin{table}[h!]
\centering
\scriptsize
\setlength{\tabcolsep}{2.4pt}
\begin{tabular}{ @{} llcccccccc @{} }
\toprule
 &  & \multicolumn{2}{c}{\textbf{BLEU} ($\uparrow$)} & \multicolumn{2}{c}{\textbf{Length Ratio} ($\to 1$)} & \multicolumn{2}{c}{\textbf{ROUGE-1} ($\uparrow$)} & \multicolumn{2}{c}{\textbf{ROUGE-2} ($\uparrow$)} \\
\cmidrule(lr){3-4} \cmidrule(lr){5-6} \cmidrule(lr){7-8} \cmidrule(lr){9-10}
\textbf{Lang} & \textbf{Model} & \textbf{Base} & \textbf{FT} & \textbf{Base} & \textbf{FT} & \textbf{Base} & \textbf{FT} & \textbf{Base} & \textbf{FT} \\
\midrule
\multirow{3}{*}{EN} & Llama3.1 & 0.053 ($\pm$ 0.001) & \textbf{0.083 ($\pm$ 0.001)} & 1.110 ($\pm$ 0.028) & 0.961 ($\pm$ 0.042) & 0.190 ($\pm$ 0.004) & \textbf{0.229 ($\pm$ 0.003)} & 0.034 ($\pm$ 0.003) & \textbf{0.057 ($\pm$ 0.001)} \\
 & Qwen3 & 0.038 ($\pm$ 0.001) & 0.081 ($\pm$ 0.002) & 1.624 ($\pm$ 0.035) & 0.933 ($\pm$ 0.057) & 0.180 ($\pm$ 0.002) & 0.220 ($\pm$ 0.001) & 0.035 ($\pm$ 0.002) & 0.054 ($\pm$ 0.002) \\
 & Ministral & 0.039 ($\pm$ 0.002) & 0.081 ($\pm$ 0.003) & 1.428 ($\pm$ 0.066) & \textbf{0.985 ($\pm$ 0.052)} & 0.186 ($\pm$ 0.003) & 0.223 ($\pm$ 0.005) & 0.032 ($\pm$ 0.003) & 0.052 ($\pm$ 0.002) \\
\midrule
\multirow{3}{*}{DE} & Llama3.1 & 0.065 ($\pm$ 0.001) & \textbf{0.095 ($\pm$ 0.002)} & 1.205 ($\pm$ 0.009) & 0.915 ($\pm$ 0.029) & 0.172 ($\pm$ 0.002) & \textbf{0.192 ($\pm$ 0.003)} & 0.041 ($\pm$ 0.001) & \textbf{0.063 ($\pm$ 0.001)} \\
 & Qwen3 & 0.049 ($\pm$ 0.001) & 0.094 ($\pm$ 0.001) & 1.633 ($\pm$ 0.019) & 0.926 ($\pm$ 0.016) & 0.171 ($\pm$ 0.002) & 0.188 ($\pm$ 0.003) & 0.040 ($\pm$ 0.001) & 0.061 ($\pm$ 0.001) \\
 & Ministral & 0.046 ($\pm$ 0.001) & 0.087 ($\pm$ 0.005) & 1.627 ($\pm$ 0.013) & \textbf{1.073 ($\pm$ 0.050)} & 0.160 ($\pm$ 0.001) & 0.182 ($\pm$ 0.003) & 0.036 ($\pm$ 0.001) & 0.059 ($\pm$ 0.002) \\
\midrule
\multirow{3}{*}{LB} & Llama3.1 & 0.007 ($\pm$ 0.001) & 0.009 ($\pm$ 0.000) & 1.291 ($\pm$ 0.026) & 0.897 ($\pm$ 0.030) & 0.113 ($\pm$ 0.002) & 0.108 ($\pm$ 0.001) & 0.012 ($\pm$ 0.001) & \textbf{0.013 ($\pm$ 0.001)} \\
 & Qwen3 & 0.003 ($\pm$ 0.000) & 0.008 ($\pm$ 0.001) & 2.427 ($\pm$ 0.036) & 0.886 ($\pm$ 0.028) & 0.079 ($\pm$ 0.001) & 0.107 ($\pm$ 0.002) & 0.008 ($\pm$ 0.000) & 0.011 ($\pm$ 0.000) \\
 & Ministral & 0.003 ($\pm$ 0.001) & \textbf{0.010 ($\pm$ 0.001)} & 2.980 ($\pm$ 0.080) & \textbf{1.077 ($\pm$ 0.038)} & 0.081 ($\pm$ 0.001) & \textbf{0.114 ($\pm$ 0.001)} & 0.008 ($\pm$ 0.000) & 0.012 ($\pm$ 0.001) \\
\bottomrule
\end{tabular}
\caption{\textbf{Extended Table for RQ1 \& RQ2: Lexical Metrics} Results show the impact of Supervised Fine-Tuning (FT) vs. prompting the base model (Base) on prediction quality. Best values per comparison unit are \textbf{bolded}. Reported values are Mean ($\pm$ Standard Deviation) across 5 independent generation runs on a hold-out test set of 650 users. All models (8B parameters) were conditioned using the combined \textit{Biography+History} strategy and trained (FT) on a dataset of 3,800 users per language.}
\end{table}

\vspace{2cm}
\begin{table}[h!]
\centering
\small
\setlength{\tabcolsep}{3pt}
\begin{tabular}{ @{} llcccccc @{} }
\toprule
 &  & \multicolumn{2}{c}{\textbf{Qwen} ($\downarrow$)} & \multicolumn{2}{c}{\textbf{Gemma} ($\downarrow$)} & \multicolumn{2}{c}{\textbf{LuxEmbedder} ($\downarrow$)} \\
\cmidrule(lr){3-4} \cmidrule(lr){5-6} \cmidrule(lr){7-8}
\textbf{Lang} & \textbf{Model} & \textbf{Base} & \textbf{FT} & \textbf{Base} & \textbf{FT} & \textbf{Base} & \textbf{FT} \\
\midrule
\multirow{3}{*}{EN} & Llama3.1 & 0.420 ($\pm$ 0.002) & \textbf{0.397 ($\pm$ 0.001)} & 0.418 ($\pm$ 0.002) & \textbf{0.402 ($\pm$ 0.001)} & 0.271 ($\pm$ 0.004) & \textbf{0.261 ($\pm$ 0.002)} \\
 & Qwen3 & 0.418 ($\pm$ 0.001) & 0.408 ($\pm$ 0.003) & 0.426 ($\pm$ 0.001) & 0.413 ($\pm$ 0.003) & 0.280 ($\pm$ 0.002) & 0.265 ($\pm$ 0.001) \\
 & Ministral & 0.424 ($\pm$ 0.004) & 0.404 ($\pm$ 0.001) & 0.428 ($\pm$ 0.004) & 0.407 ($\pm$ 0.003) & 0.283 ($\pm$ 0.004) & 0.265 ($\pm$ 0.004) \\
\midrule
\multirow{3}{*}{DE} & Llama3.1 & 0.509 ($\pm$ 0.001) & 0.504 ($\pm$ 0.002) & 0.464 ($\pm$ 0.003) & \textbf{0.455 ($\pm$ 0.005)} & 0.297 ($\pm$ 0.000) & 0.306 ($\pm$ 0.003) \\
 & Qwen3 & 0.509 ($\pm$ 0.003) & 0.512 ($\pm$ 0.006) & 0.462 ($\pm$ 0.002) & 0.466 ($\pm$ 0.005) & \textbf{0.296 ($\pm$ 0.002)} & 0.309 ($\pm$ 0.005) \\
 & Ministral & 0.505 ($\pm$ 0.002) & \textbf{0.502 ($\pm$ 0.005)} & 0.469 ($\pm$ 0.001) & 0.456 ($\pm$ 0.003) & 0.308 ($\pm$ 0.003) & 0.302 ($\pm$ 0.003) \\
\midrule
\multirow{3}{*}{LB} & Llama3.1 & 0.579 ($\pm$ 0.001) & 0.605 ($\pm$ 0.003) & 0.621 ($\pm$ 0.002) & 0.626 ($\pm$ 0.004) & \textbf{0.410 ($\pm$ 0.003)} & 0.463 ($\pm$ 0.004) \\
 & Qwen3 & \textbf{0.578 ($\pm$ 0.002)} & 0.610 ($\pm$ 0.003) & 0.622 ($\pm$ 0.003) & 0.635 ($\pm$ 0.004) & 0.415 ($\pm$ 0.002) & 0.470 ($\pm$ 0.004) \\
 & Ministral & 0.583 ($\pm$ 0.002) & 0.597 ($\pm$ 0.005) & 0.631 ($\pm$ 0.002) & \textbf{0.615 ($\pm$ 0.004)} & 0.422 ($\pm$ 0.004) & 0.443 ($\pm$ 0.003) \\
\bottomrule
\end{tabular}
\caption{\textbf{Extended Table for RQ1 \& RQ2: Embedding Distance} Results show the impact of Supervised Fine-Tuning (FT) vs. prompting the base model (Base) on embedding distance [0-2]. Best values per comparison unit are \textbf{bolded}. Reported values are Mean ($\pm$ Standard Deviation) across 5 independent generation runs on a hold-out test set of 650 users. All models (8B parameters) were conditioned using the combined \textit{Biography+History} strategy and trained (FT) on a dataset of 3,800 users per language. Embedding Models: \texttt{Qwen3-Embedding-8B} \cite{qwen3embedding}, \texttt{embeddinggemma-300m} \cite{vera2025embeddinggemma}, \texttt{LuxEmbedder} \cite{philippy-etal-2025-luxembedder}.}
\end{table}

%% file: latex/acl_latex.bbl
\begin{thebibliography}{22}
\providecommand{\natexlab}[1]{#1}

\bibitem[{Andreas(2022)}]{andreas2022language}
Jacob Andreas. 2022.
\newblock Language models as agent models.
\newblock In \emph{Findings of the Association for Computational Linguistics: EMNLP 2022}, pages 5769--5779.

\bibitem[{Dettmers et~al.(2021)Dettmers, Lewis, Shleifer, and Zettlemoyer}]{dettmers20218}
Tim Dettmers, Mike Lewis, Sam Shleifer, and Luke Zettlemoyer. 2021.
\newblock 8-bit optimizers via block-wise quantization.
\newblock \emph{arXiv preprint arXiv:2110.02861}.

\bibitem[{Durandard et~al.(2025)Durandard, Dhawan, and Poibeau}]{durandard2025llms}
No{\'e} Durandard, Saurabh Dhawan, and Thierry Poibeau. 2025.
\newblock Llms stick to the point, humans to style: Semantic and stylistic alignment in human and llm communication.
\newblock In \emph{Proceedings of the 26th Annual Meeting of the Special Interest Group on Discourse and Dialogue}, pages 206--213.

\bibitem[{Gao et~al.(2023)Gao, Lan, Lu, Mao, Piao, Wang, Jin, and Li}]{gao2023s3}
Chen Gao, Xiaochong Lan, Zhihong Lu, Jinzhu Mao, Jinghua Piao, Huandong Wang, Depeng Jin, and Yong Li. 2023.
\newblock S3: Social-network simulation system with large language model-empowered agents.
\newblock \emph{arXiv preprint arXiv:2307.14984}.

\bibitem[{Grattafiori et~al.(2024)Grattafiori, Dubey, Jauhri, Pandey, Kadian, Al-Dahle, Letman, Mathur, Schelten, Vaughan et~al.}]{grattafiori2024llama}
Aaron Grattafiori, Abhimanyu Dubey, Abhinav Jauhri, Abhinav Pandey, Abhishek Kadian, Ahmad Al-Dahle, Aiesha Letman, Akhil Mathur, Alan Schelten, Alex Vaughan, and 1 others. 2024.
\newblock The llama 3 herd of models.
\newblock \emph{arXiv preprint arXiv:2407.21783}.

\bibitem[{Hu et~al.(2025)Hu, Baumann, Lupo, Collier, Hovy, and R{\"o}ttger}]{hu2025simbench}
Tiancheng Hu, Joachim Baumann, Lorenzo Lupo, Nigel Collier, Dirk Hovy, and Paul R{\"o}ttger. 2025.
\newblock Simbench: Benchmarking the ability of large language models to simulate human behaviors.
\newblock \emph{arXiv preprint arXiv:2510.17516}.

\bibitem[{Khan et~al.(2024)Khan, Hughes, Valentine, Ruis, Sachan, Radhakrishnan, Grefenstette, Bowman, Rockt{\"a}schel, and Perez}]{khan2024debating}
Akbir Khan, John Hughes, Dan Valentine, Laura Ruis, Kshitij Sachan, Ansh Radhakrishnan, Edward Grefenstette, Samuel~R Bowman, Tim Rockt{\"a}schel, and Ethan Perez. 2024.
\newblock Debating with more persuasive llms leads to more truthful answers.
\newblock In \emph{Proceedings of the 41st International Conference on Machine Learning}, pages 23662--23733.

\bibitem[{Larooij and T{\"o}rnberg(2025)}]{larooij2025validation}
Maik Larooij and Petter T{\"o}rnberg. 2025.
\newblock Validation is the central challenge for generative social simulation: a critical review of llms in agent-based modeling.
\newblock \emph{Artificial Intelligence Review}, 59(1):15.

\bibitem[{Lin(2004)}]{lin2004rouge}
Chin-Yew Lin. 2004.
\newblock Rouge: A package for automatic evaluation of summaries.
\newblock In \emph{Text summarization branches out}, pages 74--81.

\bibitem[{Liu et~al.(2024)Liu, Chen, Zhang, Gao, Zhang, and Yan}]{liu2024skepticism}
Yuhan Liu, Xiuying Chen, Xiaoqing Zhang, Xing Gao, Ji~Zhang, and Rui Yan. 2024.
\newblock From skepticism to acceptance: simulating the attitude dynamics toward fake news.
\newblock In \emph{Proceedings of the Thirty-Third International Joint Conference on Artificial Intelligence}, pages 7886--7894.

\bibitem[{Macal and North(2009)}]{macal2009agent}
Charles~M Macal and Michael~J North. 2009.
\newblock Agent-based modeling and simulation.
\newblock In \emph{Proceedings of the 2009 winter simulation conference (WSC)}, pages 86--98. IEEE.

\bibitem[{M{\"u}nker et~al.(2025)M{\"u}nker, Schwager, and Rettinger}]{munker2025don}
Simon M{\"u}nker, Nils Schwager, and Achim Rettinger. 2025.
\newblock Don't trust generative agents to mimic communication on social networks unless you benchmarked their empirical realism.
\newblock \emph{arXiv preprint arXiv:2506.21974}.

\bibitem[{Papineni et~al.(2002)Papineni, Roukos, Ward, and Zhu}]{papineni2002bleu}
Kishore Papineni, Salim Roukos, Todd Ward, and Wei-Jing Zhu. 2002.
\newblock Bleu: a method for automatic evaluation of machine translation.
\newblock In \emph{Proceedings of the 40th annual meeting of the Association for Computational Linguistics}, pages 311--318.

\bibitem[{Philippy et~al.(2025)Philippy, Guo, Klein, and Bissyande}]{philippy-etal-2025-luxembedder}
Fred Philippy, Siwen Guo, Jacques Klein, and Tegawende Bissyande. 2025.
\newblock \href {https://aclanthology.org/2025.coling-main.753/} {{L}ux{E}mbedder: A cross-lingual approach to enhanced {L}uxembourgish sentence embeddings}.
\newblock In \emph{Proceedings of the 31st International Conference on Computational Linguistics}, pages 11369--11379, Abu Dhabi, UAE. Association for Computational Linguistics.

\bibitem[{{Qwen Team}(2025)}]{qwen3technicalreport}
{Qwen Team}. 2025.
\newblock \href {https://arxiv.org/abs/2505.09388} {Qwen3 technical report}.
\newblock \emph{Preprint}, arXiv:2505.09388.

\bibitem[{R{\"o}ttger et~al.(2024)R{\"o}ttger, Hofmann, Pyatkin, Hinck, Kirk, Sch{\"u}tze, and Hovy}]{rottger2024political}
Paul R{\"o}ttger, Valentin Hofmann, Valentina Pyatkin, Musashi Hinck, Hannah Kirk, Hinrich Sch{\"u}tze, and Dirk Hovy. 2024.
\newblock Political compass or spinning arrow? towards more meaningful evaluations for values and opinions in large language models.
\newblock In \emph{Proceedings of the 62nd Annual Meeting of the Association for Computational Linguistics (Volume 1: Long Papers)}, pages 15295--15311.

\bibitem[{Vera et~al.(2025)Vera, Dua, Zhang, Salz, Mullins, Panyam, Smoot, Naim, Zou, Chen et~al.}]{vera2025embeddinggemma}
Henrique~Schechter Vera, Sahil Dua, Biao Zhang, Daniel Salz, Ryan Mullins, Sindhu~Raghuram Panyam, Sara Smoot, Iftekhar Naim, Joe Zou, Feiyang Chen, and 1 others. 2025.
\newblock Embeddinggemma: Powerful and lightweight text representations.
\newblock \emph{arXiv preprint arXiv:2509.20354}.

\bibitem[{von Werra et~al.(2020)von Werra, Belkada, Tunstall, Beeching, Thrush, Lambert, Huang, Rasul, and Gallouédec}]{vonwerra2022trl}
Leandro von Werra, Younes Belkada, Lewis Tunstall, Edward Beeching, Tristan Thrush, Nathan Lambert, Shengyi Huang, Kashif Rasul, and Quentin Gallouédec. 2020.
\newblock Trl: Transformer reinforcement learning.
\newblock \url{https://github.com/huggingface/trl}.

\bibitem[{Wang et~al.(2025)Wang, Zhao, Ones, He, and Xu}]{wang2025evaluating}
Yilei Wang, Jiabao Zhao, Deniz~S Ones, Liang He, and Xin Xu. 2025.
\newblock Evaluating the ability of large language models to emulate personality.
\newblock \emph{Scientific Reports}, 15(1):519.

\bibitem[{Yu et~al.(2024)Yu, Zhang, Li, Fu, and Ye}]{yu2024affordable}
Yangbin Yu, Qin Zhang, Junyou Li, Qiang Fu, and Deheng Ye. 2024.
\newblock Affordable generative agents.
\newblock \emph{arXiv preprint arXiv:2402.02053}.

\bibitem[{Zhang et~al.(2025{\natexlab{a}})Zhang, Yang, Niu, Fu, Dai, and Huang}]{zhang2025spark}
Bowen Zhang, Yi~Yang, Fuqiang Niu, Xianghua Fu, Genan Dai, and Hu~Huang. 2025{\natexlab{a}}.
\newblock Spark: Simulating the co-evolution of stance and topic dynamics in online discourse with llm-based agents.
\newblock In \emph{Proceedings of the 2025 Conference on Empirical Methods in Natural Language Processing}, pages 23072--23084.

\bibitem[{Zhang et~al.(2025{\natexlab{b}})Zhang, Li, Long, Zhang, Lin, Yang, Xie, Yang, Liu, Lin, Huang, and Zhou}]{qwen3embedding}
Yanzhao Zhang, Mingxin Li, Dingkun Long, Xin Zhang, Huan Lin, Baosong Yang, Pengjun Xie, An~Yang, Dayiheng Liu, Junyang Lin, Fei Huang, and Jingren Zhou. 2025{\natexlab{b}}.
\newblock Qwen3 embedding: Advancing text embedding and reranking through foundation models.
\newblock \emph{arXiv preprint arXiv:2506.05176}.

\end{thebibliography}
